\documentclass[10pt,twocolumn,letterpaper]{article}

\usepackage{cvpr}              

\usepackage{amsmath,amsfonts,amssymb}
\usepackage{booktabs}
\usepackage{multirow}
\usepackage{algorithm}
\usepackage{algorithmic}
\usepackage{listings}
\usepackage{newfloat}
\usepackage{balance}

\definecolor{cvprblue}{rgb}{0.21,0.49,0.74}
\usepackage[pagebackref,breaklinks,colorlinks,allcolors=cvprblue]{hyperref}

\def\paperID{*****} 
\def\confName{CVPR}
\def\confYear{2026}

\title{One-shot Adaptation of Humanoid Whole-body Motion with Walking Priors}

\author{%
  Hao Huang$^{*,1,3}$, Geeta Chandra Raju Bethala$^{*,1,3}$, Shuaihang Yuan$^{1,2,3}$, Congcong Wen$^{1,3}$ \and
  Mengyu Wang$^{4}$, Anthony Tzes$^{2,3}$, Yi Fang$^{\dagger,1,2,3}$\\
  \\
  $^{1}$Embodied AI and Robotics (AIR) Lab, NYUAD\\
  $^{2}$NYUAD Center for Artificial Intelligence and Robotics (CAIR)\\
  $^{3}$New York University Abu Dhabi\\
  $^{4}$Harvard Ophthalmology AI Lab, Harvard University
}

\begin{document}

%
\twocolumn[{%
\renewcommand\twocolumn[1][]{#1}%
\maketitle
\begin{center}
\centering
\captionsetup{type=figure}
\includegraphics[width=.99\linewidth]{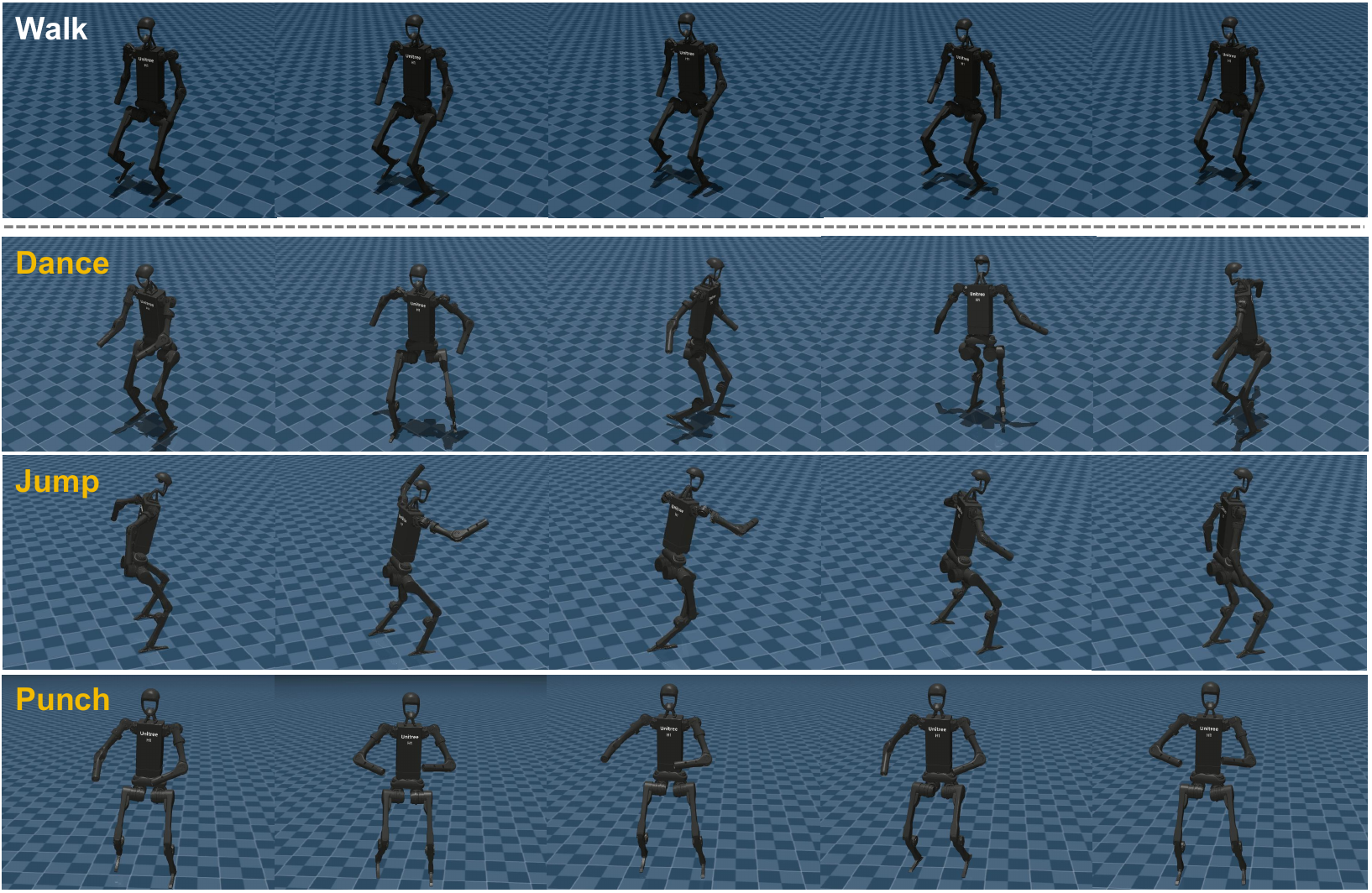}
\caption{Sampled frames from motion sequences of a humanoid (Unitree H1) performing four distinct actions in sim-to-sim transfer settings, \ie, trained in Isaac Gym and transferred to Mujoco. Walk is performed by the Base Model, while Dance, Jump, and Punch are performed by the model (Geo. FT Base Model) trained with our proposed method.}
\label{fig:teaser}
\end{center}
}]

\begingroup
\renewcommand\thefootnote{\fnsymbol{footnote}}
\footnotetext[1]{Equal contribution: hh1811, gb2643@nyu.edu}
\footnotetext[2]{Corresponding author: yfang@nyu.edu.}
\endgroup

\begin{abstract}
Whole-body humanoid motion represents a fundamental challenge in robotics, requiring balance, coordination, and adaptability to enable human-like behaviors. However, existing methods typically require multiple training samples per motion, rendering the collection of high-quality human motion datasets both labor-intensive and costly. To address this, we propose a data-efficient adaptation approach that learns a new humanoid motion from a single non-walking target sample together with auxiliary walking motions and a walking-trained base model. The core idea lies in leveraging order-preserving optimal transport to compute distances between walking and non-walking sequences, followed by interpolation along geodesics to generate new intermediate pose skeletons, which are then optimized for collision-free configurations and retargeted to the humanoid before integration into a simulated environment for policy adaptation via reinforcement learning. Experimental evaluations on the CMU MoCap dataset demonstrate that our method consistently outperforms baselines, achieving superior performance across metrics. Our code is available at: \url{https://github.com/hhuang-code/One-shot-WBM}.
\end{abstract}

\section{Introduction}
\label{sec:introduction}

\label{sec:intro}

Humanoid whole-body motion is a fundamental challenge in robotics, requiring the integration of balance, coordination, and adaptability to enable robots to act in diverse environments in a human-like manner. Progress in this area has evolved from model-based control strategies, such as trajectory optimization and feedback controllers~\cite{wensing2023optimization,elobaid2023online}, to data-driven approaches that leverage sensory feedback for real-time adjustment~\cite{long2024learning,ren2025vb,gu2025humanoid}. More recently, reinforcement learning (RL) pipelines for humanoid motion training typically rely on large datasets covering diverse human motions, such as walking, running, and jumping, with each motion category represented by multiple training samples, \ie, motion clips, to ensure generalization and robustness~\cite{cheng2024express,fu2024humanplus,he2025asap,li2025amo}. These methods retarget collected motion data to humanoid joints and train policy networks with historical proprioceptive observations, enabling the performance of agile behaviors~\cite{radosavovic2024real}.

Despite these advances, a major limitation of current training paradigms is the difficulty of collecting diverse human motion samples for a specific category with complex motion patterns. Motion capture is labor-intensive, requires specialized equipment, and often suffers from inconsistencies caused by environmental variation or performer differences~\cite{yuan2025behavior}. For example, the CMU MoCap dataset\footnote{\url{https://mocap.cs.cmu.edu/}} used in~\cite{cheng2024express,ji2025exbody2} was created by recording human subjects performing various motions in a lab equipped with 12 Vicon infrared MX-40 cameras, with subjects wearing black jumpsuits fitted with 41 reflective markers for infrared detection and triangulation of 3D marker positions. The raw data is then processed with ViconIQ software for skeleton template creation, marker labeling, and kinematic fitting, yielding over 2600 motion trials. This challenge becomes even more severe when scaling datasets for rare or complex motions, which often requires extensive teleoperation or simulation and can introduce inefficiencies and biases into policy learning~\cite{fang2024rh20t}. This raises the question of whether a humanoid whole-body motion policy can be effectively adapted to a new motion from only a single target clip, while leveraging auxiliary walking data and a pretrained walking prior, thereby reducing the need for exhaustive target-motion datasets. Although one-shot learning has been widely studied in robotic arm manipulation, where manipulators perform complex tasks from a single human demonstration~\cite{george2023one,wu2024one,zhou2025you}, its application to humanoid motion—where dynamic balance, prior locomotion competence, and multi-joint coordination introduce additional challenges—remains largely underexplored.

To address this problem, we propose a method for adapting a whole-body humanoid motion policy to a new target motion from only one target sample. A natural idea is to train a motion policy directly on this single sample, but we empirically observe that this fails for several motion categories. Instead, we first train a Base Model on multiple walking motion clips (around 130 clips), which are relatively easy to collect from Internet videos.

Then, given a single non-walking target motion clip, we generate multiple synthetic training samples to bridge the gap between the walking motions and the target motion on a pose skeleton manifold, enabling smooth interpolation across pose spaces and gradual adaptation of the walking-trained model to the target motion. Specifically, order-preserving optimal transport (OPOT)~\cite{su2017order,su2018order} is used to compute the Wasserstein distance between walking motions and the target motion, preserving temporal coherence and reducing distortion in the generated skeletons. To ensure feasibility—such as avoiding collisions among body parts—we further optimize the generated motions on the pose skeleton manifold via manifold optimization. Unlike other motion generation methods~\cite{zhang2024motiondiffuse,yu2024towards}, our approach does not require training any neural network, making it lightweight. We then fine-tune the Base Model on the generated motion data using the same strategy as in Base Model training. Experiments show that our method consistently outperforms baseline approaches across different metrics. Moreover, sim-to-sim transfer further demonstrates the robustness of the adapted model.


\section{Related Work}
\label{sec:related_work}

\label{sub:related}

\noindent \textbf{Humanoid whole-body motion.} Recent advances have substantially expanded the frontier of humanoid motion with deep RL. Transformer-based policies have shown strong performance:~\cite{radosavovic2024humanoid} formulates control as autoregressive token prediction for robust outdoor walking, while~\cite{luouniversal} builds a low-dimensional latent action space that accelerates hierarchical RL and covers 99\% of AMASS motions~\cite{mahmood2019amass}. World-model methods further improve stability and perception: HuWo~\cite{zheng2025huwo} combines a Transformer-XL~\cite{dai2019transformer} estimator with policy learning to handle complex contact dynamics, and World Model Reconstruction~\cite{sun2025learning} reconstructs implicit terrain state for blind locomotion over rough ground. Curriculum and hierarchical strategies also broaden task scope—Lin \etal~\cite{lin2025let} integrate temporal vision transformers into hierarchical RL for long-horizon trail hiking, whereas Cui \etal~\cite{cui2024adapting} combine privileged pre-training with curriculum refinement to reduce joint oscillations. Skill diversity and exploration remain active topics: Wan \etal~\cite{wan2025diversifying} augment inverse RL with quality-diversity objectives to learn gait repertoires, and Chiappa \etal~\cite{chiappa2023latent} show that lattice-structured exploration yields 18\% higher rewards on the MuJoCo Humanoid benchmark. Complementary work emphasizes data efficiency and imitation learning: CROSSLOCO~\cite{li2023crossloco} drives bipedal agents with cross-modal motion priors, while Dugar \etal~\cite{dugar2024learning} track arbitrary pose subsets for robust whole-body imitation. Further progress addresses expressive whole-body control: ExBody~\cite{cheng2024express} and Exbody2~\cite{ji2025exbody2} use the CMU MoCap motion repertoire for motion mimicking; HumanPlus~\cite{fu2024humanplus} studies real-time mirroring of human subjects; ASAP~\cite{he2025asap} improves hardware transfer; and H2O~\cite{he2024learning} enables low-latency kinesthetic control. Humanoid-X~\cite{mao2025universal} converts 20 million human-video pose--text pairs into a large humanoid model for scalable text-driven whole-body control. In contrast, we study adaptation to a new humanoid motion from a single target clip while leveraging auxiliary walking data and a walking prior, reducing the burden of target-motion data collection.

\noindent \textbf{Human motion generation.} Recent work on human motion generation has explored VAE~\cite{kingma2013auto}, diffusion~\cite{ho2020denoising}, and hybrid models built on large motion corpora. Tevet \etal~\cite{tevet2023human} pioneer text- and action-conditioned denoising diffusion for long-range sequences, and Motion Latent Diffusion~\cite{chen2023executing} accelerates this approach by operating in a learned VAE latent space. Physics awareness is introduced by PhysDiff~\cite{yuan2023physdiff}, which projects each denoised step through a simulator to suppress foot-sliding and penetration, and by BioMoDiffuse~\cite{kang2025biomodiffuse}, which embeds Euler--Lagrange dynamics to ensure biomechanical plausibility. Controllability advances include Guided Motion Diffusion~\cite{karunratanakul2023guided}, which optimizes motions toward goal functions during sampling, and SMooDi~\cite{zhong2024smoodi}, which enables arbitrary motion-style transfer via multi-condition diffusion. Goal-directed generation is addressed by WANDR~\cite{diomataris2024wandr}, which uses a feedback conditional VAE to reach 3D targets without RL, whereas CrossDiff$^3$~\cite{ren2024realistic} leverages cross-domain diffusion to couple 2D and 3D constraints. Broader environmental coverage is pursued by TRAM~\cite{wang2024tram} for long-range path-plus-pose synthesis and by PC-MRL~\cite{mo2024motion}, which homogenizes diverse point-cloud motions for cross-dataset interpolation. Collectively, these works establish a strong foundation of data-driven priors, physics-based constraints, and controllable diffusion processes. Zhao \etal~\cite{zhao2024denoising} propose a conditional diffusion model with bidirectional Markov chains to generate realistic, diverse, variable-length 3D skeleton motions for specified action classes, while SALAD~\cite{hong2025salad} explicitly captures relationships among skeletal joints, temporal frames, and text descriptors. Unlike prior studies, our method generates novel human motions without training any neural network, making it lightweight and well suited for whole-body robot motion policy learning.


\section{Methods}
\label{sec:methods}

\label{sub:method}
\begin{figure*}[!htb]
\centering
\includegraphics[width=.99\linewidth]{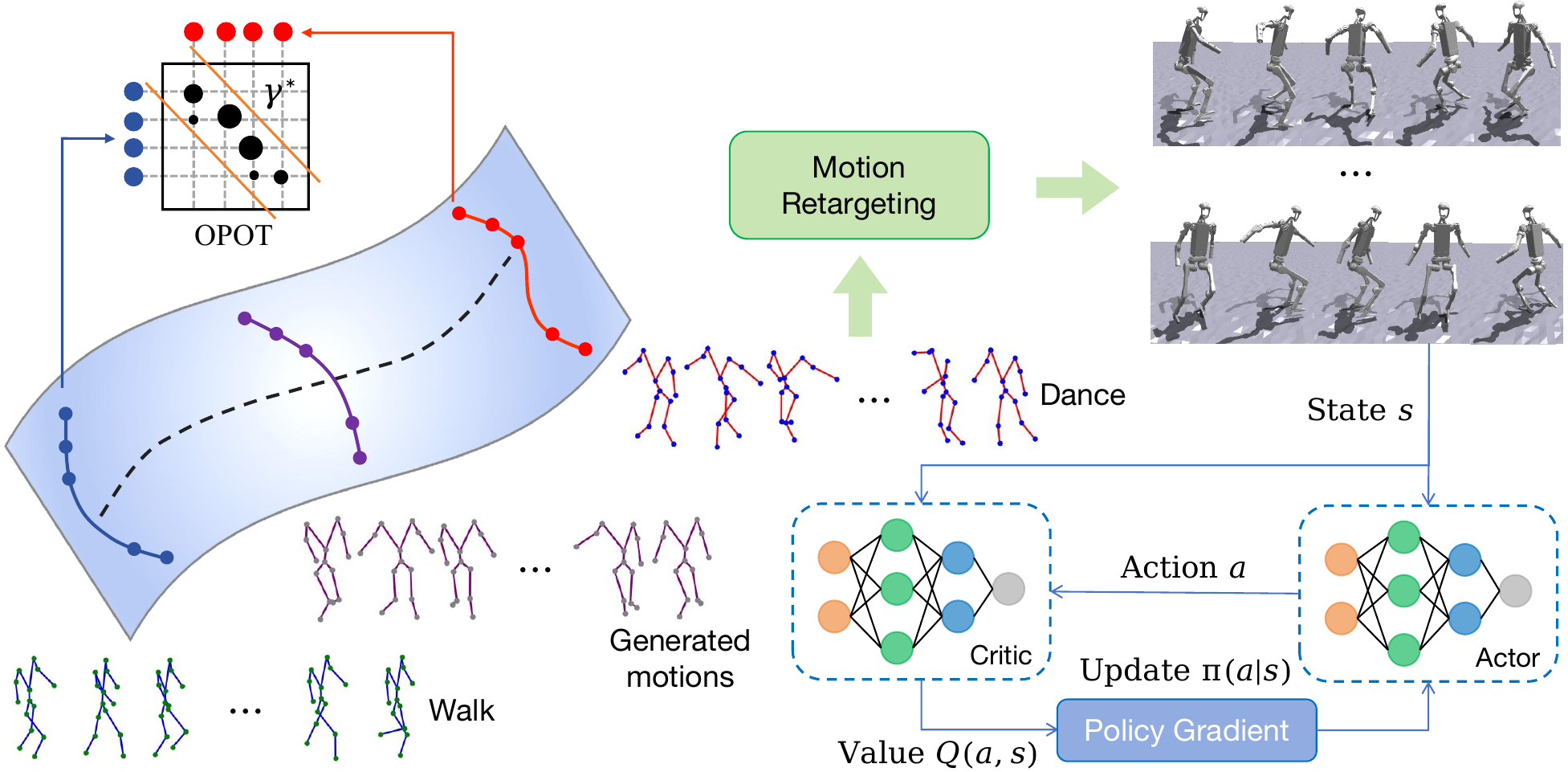}
\caption{Given a sequence of walking motion pose skeletons and a target sequence comprising non-walking motions, we employ order-preserving optimal transport (OPOT) to compute the distance between these two sequences. Subsequently, we interpolate and sample novel pose skeletons along the geodesics connecting the walking and non-walking sequences. These sampled skeletons, together with the target sequence, are then retargeted to a humanoid robot and integrated into a simulated environment for training a whole-body motion policy via the Proximal Policy Optimization (PPO) algorithm.}
\label{fig:pipeline}
\end{figure*}

A common framework in recent work~\cite{he2024learning,cheng2024express,ji2025exbody2} for training humanoids to perform diverse and expressive whole-body motions begins by curating human motion clips from datasets such as CMU MoCap or AMASS~\cite{mahmood2019amass} using video-to-pose estimation models or visual tracking systems. These clips are then retargeted to the robot's kinematic structure (\eg, Unitree H1 with 19 DoFs) by mapping 3D human joint rotations to humanoid joint angles. From the retargeted clips, behavior goals (upper-body joint positions and keypoints) and root movement goals (linear velocity, roll/pitch/yaw, and height) are extracted. A goal-conditioned reinforcement learning policy is then trained with PPO~\cite{schulman2017proximal} in a simulated environment (\ie, Isaac Gym\footnote{\url{https://developer.nvidia.com/isaac-gym}}) with randomized terrains and carefully designed rewards, where the upper body imitates reference motions via tracking rewards, while the legs follow root commands to maintain balance without strict imitation. An overview of this framework is shown in the right part of Figure~\ref{fig:pipeline}. However, nearly all existing methods require many motion clips from the same category to learn a behavior successfully. Collecting large numbers of clips for diverse complex motions, whether from videos or teleoperation, is labor-intensive and tedious. By contrast, walking videos—the most common motion in daily life—are easy to obtain on the Internet, whereas other diverse motion videos are much harder to acquire. Motivated by this, we explore a solution that adapts an expressive whole-body humanoid motion model to a new target behavior using a single non-walking target clip together with multiple easy-to-acquire walking videos. The core innovation is to generate multiple intermediate motions that bridge walking and target non-walking motions, thereby enabling smooth transfer from the walking prior during policy adaptation.

%

\subsection{Geodesic Distance for Pose Skeletons}
\label{subsec:pose_geo}
First, we define a mathematical representation for human pose skeletons extracted from human motion clips. Formally, a \emph{pose skeleton} at each timestep is represented as $(x, \{q_1, \dots, q_J\})$~\cite{vemulapalli2014human,vemulapalli2016rolling}, where:
\begin{itemize}
    \item $x \in \mathbb{R}^3$ is a translation vector, specifying the position of the skeleton's root (pelvis) in 3D Euclidean space.
    \item $\{q_1, \dots, q_J\}$ is a list of $J$ unit quaternions, where each $q_j \in \mathbb{H}_1 \subset \mathbb{R}^4$ represents the rotation of the $j$-th joint. Here, $\mathbb{H}_1 = \{ q \in \mathbb{R}^4 : \|q\|_2 = 1 \}$ denotes the unit quaternion group, which is a double cover of $\text{SO}(3)$.
\end{itemize}
One benefit of this representation is that it is invariant \wrt skeleton scale. The \emph{geodesic distance} between two poses, $(x_1, \{q_{1,1}, \dots, q_{1,J}\})$ and $(x_2, \{q_{2,1}, \dots, q_{2,J}\})$, is a metric that combines:
\begin{itemize}
    \item The Euclidean distance between root translations in $\mathbb{R}^3$:
    \begin{equation}
    \small
    d_{\text{t}}(x_1, x_2) = \|x_1 - x_2\|_2 \enspace.
    \end{equation}
    \item The sum of geodesic distances on $\text{SO}(3)$ for all the corresponding joint rotations, each of which is computed via:
    \begin{equation}
    \small
        d_{\text{r}}(q_{1,j}, q_{2,j}) = 2 \arccos(|\langle q_{1,j}, q_{2,j} \rangle|)\enspace,
    \end{equation}
    where $\langle q_{1,j}, q_{2,j} \rangle = q_{1,j} \cdot q_{2,j}$ is the Euclidean dot product in $\mathbb{R}^4$, and the absolute value ensures the shortest arc (since $q$ and $-q$ represent the same rotation in $\text{SO}(3)$).
\end{itemize}

Since the pose representation space is the direct product of Lie groups $\mathbb{R}^3 \times \mathrm{SO}(3)^J$, allowing the metric to be defined as the weighted sum of the independent geodesic distances on the translational component in $\mathbb{R}^3$ and the rotational components in each $\mathrm{SO}(3)$. Thus, the total geodesic distance is a weighted combination:
\begin{equation}
\small
d((x_1, \{q_{1,j}\}), (x_2, \{q_{2,j}\})) = d_{\text{t}}(x_1, x_2) + w \sum_{j=1}^N d_{\text{r}}(q_{1,j}, q_{2,j})\enspace,
\label{eq:frame_dist}
\end{equation}
where $w \in \mathbb{R}_{\geq 0}$ is a weight parameter balancing the contribution of translational and rotational components. In our settings, we set $w=1$ for all cases. With this definition, we can measure the distance between two pose skeletons at different time steps across different motion clips. Then, we can further compute the discrepancy between the walking and non-walking motion sequences, as detailed below.

\subsection{Order-Preserving Optimal Transport for Skeleton Sequence Alignment}
\label{subsec:opw}
Given a skeleton sequence from a walking clip, denoted as $S = \{s_1, \dots, s_N\}$ where $s_n = (x_n, \{q_{n,1}, \dots q_{n, J}\})$ is the pose skeleton at the $n$th timestep as defined above. Similarly, a skeleton sequence from a non-walking target motion clip is denoted as $T=\{t_1, \dots, t_M\}$ where $t_m = (x_m, \{q_{m,1}, \dots q_{m, J}\})$. We define a \textit{distance} to measure the discrepancy between these two skeleton sequences. Optimal transport with Wasserstein distance~\cite{villani2008optimal} is a natural choice. However, classical optimal transport treats pose skeletons at each timestep as unordered sets, allowing non-monotonic mappings that violate the inherent chronological structure of motion sequences.

Order-Preserving Wasserstein (OPW) distance~\cite{su2017order,su2018order} between two sequences handling local temporal distortions and periodic sequences, extends the classical Wasserstein distance by incorporating temporal regularizations to preserve order in sequential data. Specifically, the OPW distance views these sequences as empirical probability measures with uniform masses: \(\alpha = \mathbf{1}_N / N \in \mathbb{R}^N\) and \(\beta = \mathbf{1}_M / M \in \mathbb{R}^M\), where \(\mathbf{1}_k\) denotes the all-ones vector of length \(k\). The cost matrix \(D \in \mathbb{R}^{N \times M}\) is computed for each pair of pose skeletons as defined in Eq.~\ref{eq:frame_dist}, \ie, $D(n, m) = d((x_n, \{q_{n,j}\}), (x_m, \{q_{m,j}\}))$. The OPW additionally incorporates two regularizations:
\begin{itemize}
    \item Inverse Difference Moment Regularization: Encourages local homogeneous transport by favoring couplings near the ``diagonal'' in normalized indices. Defined via the matrix \(H \in \mathbb{R}^{N \times M}\) where:
    \begin{equation}
    \small
    H_{nm} = \frac{\lambda_1}{(n/N - m/M)^2 + 1} \enspace,
    \end{equation}
    with \(\lambda_1 > 0\) as the weight parameter.
    \item KL-Divergence Regularization with Gaussian Prior: Aligns the transport plan with a prior distribution \(P \in \mathbb{R}^{N \times M}\) modeling expected couplings based on positional similarity. The prior is a scaled Gaussian:
    \begin{equation}
    \small
    P_{nm} = \frac{1}{\delta \sqrt{2\pi}} \exp\left( -\frac{d_{nm}^2}{2 \delta^2} \right) \enspace,
    \end{equation}
    where \(d_{nm} = \left| \frac{n}{N} - \frac{m}{M} \right| / \sqrt{\frac{1}{N^2} + \frac{1}{M^2}}\), and \(\delta > 0\) is the standard deviation parameter.
\end{itemize}

Then, the OPW distance is formulated as the minimum cost under these regularizations, approximated via entropic regularization with parameter \(\lambda_2 > 0\):
\begin{equation}
\small
d^{\lambda_1, \lambda_2}(S, T) = \min_{\Gamma \in U(\alpha, \beta)} \langle \Gamma, D \rangle - \lambda_1 \text{I}(\Gamma) + \lambda_2 \text{KL}(\Gamma \| P)\enspace,
\label{eq:opw_dist}
\end{equation}
where \(U(\alpha, \beta)\) is the transport polytope for alignment, \(\langle \cdot, \cdot \rangle\) is the Frobenius inner product, \(I(\Gamma) = \sum_{nm} \Gamma_{nm} H_{nm} / \lambda_1\) (adjusted for scaling), and \(\text{KL}(\Gamma \| P)\) is the Kullback-Leibler divergence. Eq.~\ref{eq:opw_dist} is solved using Sinkhorn iterations~\cite{su2018order}. In our settings, we set $\lambda_1=50$, $\lambda_2=0.1$, $\delta=1$, and the maximum number of iterations to 20 for all cases. After solving Eq.~\ref{eq:opw_dist}, we can get 1. the optimal per-frame \textit{alignment} (a.k.a transport plan) between the walking and non-walking motion sequences, and 2. the shortest \textit{distance} between these two sequences. This process is shown in the left part of Figure~\ref{fig:pipeline}. Then, we select $Q=10$ walking motions with the smallest shortest distances \wrt the target non-walking motions to generate new intermediate motions.

\subsection{Soft Transport Plans to Hard Assignments}
\label{subsec:hard_plan}
The solution to Eq.~\ref{eq:opw_dist} is a soft transport plan, \ie, the pose skeleton at each timestep in the non-walking target sequence can be possibly aligned to more than one pose skeleton across multiple (usually two to three) adjacent timesteps in the walking sequence. However, our proposed motion generation approach necessitates a hard assignment of each target pose skeleton at every timestep to exactly one walking pose skeleton, permitting reuse of the latter. Thus, we propose a method for projecting a soft transport plan onto a hard assignment matrix, ensuring each target is matched to exactly one source, with sources potentially reused. 
The method leverages the linear sum assignment problem, solved using the Hungarian algorithm, to maximize the total assignment score derived from the input soft transport plan.

Let \(\Gamma^\ast \in \mathbb{R}^{n \times m}_{\geq 0}\) be the optimal soft transport plan derived from Eq.~\ref{eq:opw_dist}. The matrix \(\Gamma^\ast\) represents a probabilistic coupling, where \(\Gamma_{nm} \geq 0\) indicates the strength of association between source \(s_n\) and target \(t_m\). The goal is to produce a hard assignment matrix \(\gamma \in \{0, 1\}^{n \times m}\), where \(\gamma_{nm} = 1\) if source \(s_n\) is assigned to target \(t_m\), and \(\gamma_{nm} = 0\) otherwise, subject to the constraint that each target skeleton is assigned exactly one walking skeleton:
\begin{equation}
\small
\sum_{n=1}^N \gamma_{nm} = 1, \quad \forall m \in \{1, \dots, M\},
\end{equation}
while allowing sources to be reused (\ie, \(\sum_{m=1}^M \gamma_{nm} \geq 0\)). The problem is formulated as maximizing the total assignment score:
\begin{equation}
\small
\max_{\gamma \in \mathcal{T}} \sum_{n=1}^N \sum_{m=1}^M \gamma_{nm} \Gamma^\ast_{nm}\enspace,
\end{equation}
where \(\mathcal{T} = \{ \gamma \in \{0, 1\}^{N \times M} \mid \sum_{n=1}^N \gamma_{nm} = 1, \forall m \}\) is the set of valid assignment matrices. To use standard optimization algorithms, we take \(C = -\Gamma^\ast\), transforming the maximization into a minimization:
\begin{equation}
\small
\min_{\gamma \in \mathcal{T}} \sum_{n=1}^N \sum_{m=1}^M C_{nm} \gamma_{nm}\enspace.
\end{equation}

Note that we need to take two cases into consideration based on the relative sizes of \(N\) and \(M\):

\textbf{Case 1: sufficient walking skeletons (\(N \geq M\))}: When the walking clip is longer than or equal to the target motion clip, each target pose skeleton can be assigned a unique walking one. The cost matrix \(C = -\Gamma^\ast \in \mathbb{R}^{N \times M}\) is transposed to shape \((M, N)\) to align with the convention of assigning targets to sources. The Hungarian algorithm solves:
\begin{equation}
\small
\begin{aligned}
    & \min \sum_{m=1}^M \sum_{n=1}^N C_{nm} \tilde{\gamma}_{mn}, \enspace \\
    & \text{s.t.} \enspace \sum_{n=1}^N \tilde{\gamma}_{mn} = 1, \forall m, \enspace \sum_{m=1}^M \tilde{\gamma}_{mn} \leq 1, \forall n,
\end{aligned}
\end{equation}
where \(\tilde{\gamma} \in \{0, 1\}^{m \times n}\) assigns each target skeleton to a walking skeleton. The algorithm returns indices \((m, n)\) such that \(\tilde{\gamma}_{mn} = 1\). These are mapped to \(\gamma \in \{0, 1\}^{n \times m}\) by setting \(\gamma_{nm} = 1\) if \(\tilde{\gamma}_{mn} = 1\), ensuring each target skeleton is assigned exactly one walking skeleton.

\textbf{Case 2: insufficient walking skeletons (\(N < M\))}: When the walking clip is shorter than the target motion clip, walking skeletons must be reused. To achieve this, the cost matrix $C$ is tiled to create virtual skeletons. The number of repetitions is computed as $r = \left\lceil \frac{M}{N} \right\rceil$, and the cost matrix is replicated \(r\) times along the row dimension, yielding a tiled matrix \(C_{\text{tiled}} \in \mathbb{R}^{rN \times M}\). The first \(M\) rows are selected to form a square matrix of shape \((M, M)\). The Hungarian algorithm is applied to the transposition of this matrix to solve a one-to-one assignment problem:
\begin{equation}
\small
\begin{aligned}
    & \min \sum_{m=1}^M \sum_{n=1}^M C_{\text{tiled}, nm} \tilde{\gamma}_{mn}, \enspace \\
    & \text{s.t.} \enspace \sum_{n=1}^M \tilde{\gamma}_{mn} = 1, \forall m, \enspace \sum_{m=1}^M \tilde{\gamma}_{mn} = 1, \forall n\enspace.
\end{aligned}
\end{equation}
The resulting assignments map each target skeleton \(t_m\) to a virtual walking skeleton \(s_{n_{\text{virtual}}}\). The real walking skeleton index is computed as \(n_{\text{real}} = (n_{\text{virtual}} \mod N)\), and the assignment is recorded in \(\gamma\) by setting \(\gamma_{n_{\text{real}}, m} = 1\).

The above process returns \(\gamma \in \{0,1\}^{N\times M}\), where each column sums to 1, ensuring every target non-walking skeleton is assigned exactly one walking skeleton. Then, we can generate new pose skeletons along the geodesics connecting each paired walking and non-walking target motions at every timestep.

\subsection{Pose Skeletons Sampling along Geodesics}
\label{subsec:sampling}
When each non-walking target pose skeleton is matched to a walking skeleton, we can interpolate and sample poses along the geodesic path in the product manifold $\mathbb{R}^3 \times \text{SO}(3)^J$ to generate new pose skeletons. On this manifold, $\mathbb{R}^3$ is equipped with the Euclidean metric, and each $\text{SO}(3)$ component uses the geodesic metric induced by the unit quaternion representation. 
The \emph{geodesic path} between two paired poses $s_n = (x_n, \{q_{n,j}\}_{j=1}^J)$ and $t_m = (x_m, \{q_{m,j}\}_{j=1}^J)$ is the shortest path in this manifold, parameterized by $\tau \in [0, 1]$. The interpolated pose at $\tau$ is:
\begin{equation}
\small
p(\tau) = (x(\tau), \{q_j(\tau)\}_{j=1}^J)\enspace,
\end{equation}
where:
\begin{itemize}
    \item $x(\tau) = (1 - \tau) x_n + \tau x_m$ is the linear interpolation in $\mathbb{R}^3$ for the root translation.
    \item $q_j(\tau) = \text{SLERP}(q_{n,j}, q_{m,j}, \tau)$ is the spherical linear interpolation on $\mathbb{H}_1$ for the $j$th joints' rotation, defined for unit quaternions $q_{n,j}, q_{m,j}$ as:
    \begin{equation}
    \small
    q_j(\tau) = \frac{\sin((1-\tau)\theta_j)}{\sin(\theta_j)} q_{n,j} + \frac{\sin(\tau\theta_j)}{\sin(\theta_j)} q_{m,j}\enspace,
    \end{equation}
    where $\theta_j = \arccos(\langle q_{n,j}, q_{m,j} \rangle)$ is the angle between quaternions, and $\langle \cdot, \cdot \rangle$ is the Euclidean dot product in $\mathbb{R}^4$. 
    For antipodal quaternions ($q_{m,j} \approx -q_{n,j}$), a shortest-path adjustment is typically applied.
\end{itemize}
Through interpolation and sampling, we generated new pose skeletons that exhibit motion intermediate between walking and the target motions. For each of the selected $Q = 10$ walking motions, we generate six new pose skeletons along geodesics.

\subsection{Optimization for Collision-Free Skeletal Poses}
\label{subsec:opt}
The generated poses above may cause bone collisions, so we introduce an optimization routine to make them collision-free for robot retargeting. Given an interpolated pose $p(\tau)$, we first apply forward kinematics to compute the world-space transforms and positions of all joints by composing local transformations along the kinematic tree, yielding a $J \times 3$ matrix of joint positions in the world frame.

For collision checking, each joint is modeled as a sphere and each bone as a capsule. A fast batched segment--segment distance kernel implements the collision detection algorithm of~\cite{ericson2004real}. Based on these distances, we define two differentiable energies: a sphere energy that penalizes interpenetration between non-adjacent joint spheres while masking parent/child and grandparent relations, and a capsule energy that penalizes overlaps between non-adjacent bones by modeling each bone as a capsule (with radius inherited from its parent joint) and applying a hinge-style penalty when two capsules penetrate each other.
The sphere energy sums squared penetration depths over non-adjacent joint pairs, excluding parents, children, and grandparents to focus on true self-collisions. The capsule energy extends this to bones by gathering bone segments, computing all-pairs segment distances, and summing squared penetrations over non-adjacent bone pairs after masking those that share joints.
Because forward kinematics and both collision energies are differentiable \wrt $p(\tau)$, we optimize the pose, as outlined in Algorithm~\ref{alg:opt}, by minimizing the combined sphere and capsule energies with Riemannian gradient descent on the product manifold $\text{SO}(3)^J$ for joint rotations. This ensures descent along geodesic directions on the rotation manifold, with learning rate $lr$ controlling the step size. More details for collision detection is provided in~\cite{huang2026one}.
\begin{algorithm}
\caption{Pose Collision-Free Optimization Algorithm}
\begin{algorithmic}[1]
\STATE $\mathbf{q} \gets \text{clone}(p(\tau))$ with gradients enabled
\FOR{each step up to maximum number}
    \STATE $\mathbf{x} \gets \text{FK}(\mathbf{q})$ \COMMENT{\textcolor{blue}{forward kinematics}}
    \STATE $E \gets \text{sphere\_energy}(\mathbf{x}) + \cdot \text{capsule\_energy}(\mathbf{x})$
    \IF{$E < 10^{-6}$}
        \STATE Break \COMMENT{\textcolor{blue}{terminate early if energy is small}}
    \ENDIF
    \STATE $\nabla_{\mathbf{q}} E \gets \text{backward}(E)$ \COMMENT{\textcolor{blue}{back-propagate}}
    \STATE $\mathbf{g} \gets \nabla_{\mathbf{q}} E$
    \STATE $\mathbf{g}_{\text{proj}} = \mathbf{g} - (\mathbf{q} \cdot \mathbf{g}) \mathbf{q}$ \COMMENT{\textcolor{blue}{project gradient; $\cdot$ is dot product}}
    \STATE $\mathbf{q} \gets \mathbf{q} - lr \cdot \mathbf{g}_{\text{proj}}$ \COMMENT{\textcolor{blue}{$lr$ is learning rate}}
    \STATE $\mathbf{q} \gets \text{quat\_normalize}(\mathbf{q})$ \COMMENT{\textcolor{blue}{project back to unit sphere}}
    \STATE Zero $\mathbf{q}$.grad
\ENDFOR
\STATE $\mathbf{x}_{\text{final}} \gets \text{FK}(\text{quat\_normalize}(\mathbf{q}))$ \COMMENT{\textcolor{blue}{$\mathbf{x}_{\text{final}}$ is used for motion policy training}}
\end{algorithmic}
\label{alg:opt}
\end{algorithm}
%


\section{Experiments}
\label{sec:experiments}

\label{sec:exp}

\noindent \textbf{Human motion data curation.} The CMU MoCap dataset contains 780 human motion clips, containing not only walking, but also other expressive motion behaviors. Unlike previous works~\cite{cheng2024express,ji2025exbody2} which train a model with a large portion of MoCap data, we formulate our training scheme in two phases. First, we train a Base Model on multiple walking motions, and then fine-tune the Base Model with our generated motions along the geodesics connecting the walking motions and the single target motion. Note that the motion clips for training and evaluation are performed by different actors to increase the challenges of motion variability. The select motions and the number of evaluation clips are listed in Table~\ref{tab:data}.

\noindent \textbf{Baselines and evaluation metrics.} Our baselines include:
\begin{itemize}
    \item Base Model: We train this model using the 135 Walk clips from scratch.
    \item Single Motion Motion: For each category, we train a model using only one training clip from scratch.
    \item Single Finetune (FT) Base Model: For each category, we finetune the Base Model using only one training clip.
    \item Geodesic (Geo.) Motion Base Model: For each category, we train a model using the motions sampled along the geodesics from scratch.
    \item Geodesic (Geo.) Finetune (FT) Base Model: For each category, we finetune the Base Model using the motions sampled along the geodesics.
\end{itemize}

Following~\cite{cheng2024express}, we use the metrics below:
\begin{itemize}
    \item Mean Episode Linear Velocity Tracking Reward (MELV).
    \item Mean episode roll pitch tracking reward (MERP).
    \item Mean episode key body tracking reward (MEK).
    \item Mean episode lengths (MEL).
\end{itemize}
\begin{table}[ht]
\centering
\small
\renewcommand{\arraystretch}{1.1} 
\begin{tabular}{l|cc|cc}
\toprule
\multirow{2}{*}{\textbf{Categories}} & \multicolumn{2}{c|}{\textbf{Training}} & \multicolumn{2}{c}{\textbf{Evaluation}} \\ 
\cmidrule{2-5} 
                          & Clips       & Duration (s)      & Clips        & Duration (s)       \\ 
\midrule
Walk                      & 135         & 1752.5          & -           & -                \\
\midrule
Basketball                & 1           & 50.5            & 13          & 135.1            \\
Dance                     & 1           & 7.7             & 74          & 1455.9           \\
Jump                      & 1           & 3.5             & 23          & 84.9             \\
Punch                     & 1           & 15.4            & 12          & 316.6            \\
Wash                      & 1           & 14.2            & 8           & 250.1            \\ 
\bottomrule
\end{tabular}
\caption{Statistics for training and evaluation motion clips.}
\label{tab:data}
\end{table}
%

\subsection{Results and Analysis}
\label{subsec:result}

\begin{table*}[!ht]
\centering
\small
\renewcommand{\arraystretch}{1.1} 
\setlength{\tabcolsep}{9pt}
\begin{tabular}{l|cccc|cccc}
\toprule
\multirow{2}{*}{\textbf{Methods}}   & \multicolumn{4}{c|}{\textbf{Basketball}}              & \multicolumn{4}{c}{\textbf{Dance}}           \\ 
\cmidrule{2-9} 
                           & MELV   & MERP   & MEKB   & MLE               & MELV   & MERP   & MEKB   & MLE      \\ 
\midrule
Base Model                 & 17.525 & 43.452 & 37.542 & \textbf{1501.990}          & 4.286  & 19.956 & 15.146 & 789.447  \\
Single Motion Model              & \underline{17.657} & 41.408 & 40.220 & \textbf{1501.990} & 3.803  & \underline{29.389} & \underline{22.132} & \underline{1220.065} \\
Single FT Base Model              & \textbf{20.143} & 43.247 & \underline{41.347} & \textbf{1501.990} & 6.323  & 25.625 & 19.792 & 960.378  \\
\midrule
Geo. Motion Model & 14.206 & \textbf{45.458} & \textbf{41.618} & \textbf{1501.990} & \underline{7.209}  & 22.539 & 16.318 & 862.318  \\
Geo. FT Base Model  & 15.670 & \underline{45.449} & 39.218 & \textbf{1501.990} & \textbf{14.133} & \textbf{36.744} & \textbf{26.563} & \textbf{1421.610} \\ 
\bottomrule
\end{tabular}
\caption{Quantitative evaluation of motion models on Basketball and Dance using MELV, MERP, MEKB, and MLE metrics.}
\label{tab:basketball_dance}
\end{table*}
\begin{table*}[!ht]
\centering
\small
\renewcommand{\arraystretch}{1.1} 
\setlength{\tabcolsep}{4.5pt}
\begin{tabular}{l|cccc|cccc}
\toprule
\multirow{2}{*}{\textbf{Methods}}   & \multicolumn{4}{c|}{\textbf{Jump}}                    & \multicolumn{4}{c}{\textbf{Punch}}           \\ 
\cmidrule{2-9} 
                           & MELV   & MERP   & MEKB   & MLE               & MELV   & MERP   & MEKB   & MLE      \\ 
\midrule
Base Model                 & 8.464  & \textbf{42.959} & \textbf{33.048} & \textbf{1487.360}          & 8.405  & 38.967 & 31.275 & 1256.202 \\
Single Motion Model              & 1.202$_{(\times)}$  & 4.657$_{(\times)}$  & 2.500$_{(\times)}$ & 163.030$_{(\times)}$  & 36.122$_{(\times)}$ & 36.752$_{(\times)}$ & 16.159$_{(\times)}$ & 1459.245$_{(\times)}$ \\
Single FT Base Model              & 6.686  & 37.217 & 27.179 & 1252.151 & \textbf{22.559} & 44.435 & 35.164 & \textbf{1501.990} \\
\midrule
Geo. Motion Model & \underline{10.869} & 41.927 & 30.062 & 1456.650 & 15.218 & \underline{44.658} & \underline{35.795} & \textbf{1501.990} \\
Geo. FT Base Model  & \textbf{14.294} & \underline{42.063} & \underline{32.865} & \underline{1458.304} & \underline{17.904} & \textbf{46.369} & \textbf{36.375} & \textbf{1501.990} \\ 
\bottomrule
\end{tabular}
\caption{Quantitative evaluation of motion models on Jump and Punch using MELV, MERP, MEKB, and MLE metrics.}
\label{tab:jump_punch}
\end{table*}
\begin{table}[!ht]
\centering
\small
\renewcommand{\arraystretch}{1.1} 
\setlength{\tabcolsep}{1.5pt}
\begin{tabular}{l|cccc}
\toprule
\multirow{2}{*}{\textbf{Methods}}   & \multicolumn{4}{c}{\textbf{Wash}}                     \\ 
\cmidrule{2-5} 
                           & MELV   & MERP   & MEKB   & MLE               \\ 
\midrule
Base Model                 & 5.864  & 44.144 & 29.704 & \textbf{1501.990}          \\
Single Motion Model              & 16.862$_{(\times)}$ & 18.915$_{(\times)}$ & 9.684$_{(\times)}$ & 720.329$_{(\times)}$  \\
Single FT Base Model              & 7.735  & \underline{44.651} & \textbf{34.709} & \textbf{1501.990} \\
\midrule
Geo. Motion Model & \underline{12.336} & 44.192 & \underline{32.591} & \textbf{1501.990} \\
Geo. FT Base Model  & \textbf{13.187} & \textbf{44.902} & 32.190 & \textbf{1501.990} \\ 
\bottomrule
\end{tabular}
\caption{Quantitative evaluation of motion models on Wash using MELV, MERP, MEKB, and MLE metrics.}
\label{tab:wash}
\end{table}

The results for five target non-walking motions are shown in Table~\ref{tab:basketball_dance} to Table~\ref{tab:wash}. Across all five motions, two clear trends emerge. First, relying on a single non-walking clip without prior knowledge (Single Motion Model) is risky: the robot falls in three of the five tasks ($\times$ marks), even though it achieves higher scores in some cases. Starting from the Base Model and fine-tuning on that same single clip (Single FT Base Model) already improves stability and often yields competitive second-tier results. However, the most consistent gains come from training with motions generated along geodesics. Training only on these synthetic sequences (Geo. Motion Model) gives the best or second-best results on every metric for Basketball and brings solid improvements for Jump, Punch, and Wash. When geodesic data are used to fine-tune the Base Model (Geo. FT Base Model), the method performs best overall, attaining the top or second-best scores in 17 of 20 metrics. Second, this pattern is fairly consistent across metrics. Geo. FT Base Model achieves the highest MELV and MEKB values for four motions, and the highest MERP for Dance, Punch, and Wash, while also delivering competitive results and the best performance for Basketball and Jump. Overall, training from scratch or fine-tuning the Base Model with our interpolated motions is the most reliable route to strong quantitative performance and robust execution, whereas naively training from scratch on a single motion clip is prone to failure.


\section{Ablation Study}
\label{sec:ablation_study}

\label{sec:abl}

\noindent \textbf{Effect of the transport formulation.} We replace order-preserving optimal transport with classical, permutation-free optimal transport on the Jump motion, and report the results in Table~\ref{tab:abl}. We observe consistent performance drops across all four metrics, indicating that the matching strategy plays an important role in downstream policy learning. Because the order-preserving formulation explicitly respects the temporal structure of the walking sequence, it produces correspondences that remain coherent over time and avoid unrealistic frame-to-frame reassignments. As a result, it yields geodesic trajectories whose kinematic statistics (\eg, phase progression and velocity profiles) align more closely with the target jump sequence, leading to smoother intermediate motions.

\noindent \textbf{Effect of the number of geodesic samples.} We keep the order-preserving optimal transport fixed and vary the number of interpolated poses used for training. Performance improves monotonically as the number of samples increases from one to three to six, with six achieving the best scores overall. This trend suggests that the amount of synthesized supervision is important in our single-target adaptation setting, and that overly sparse interpolation can leave sizable gaps between the source and target motion distributions. Denser sampling along the geodesic gives the policy richer coverage of intermediate poses and transition dynamics. Consequently, the learned behaviors exhibit smoother trajectories and more accurate end-point kinematics.

\begin{table}[!ht]
\centering
\small
\renewcommand{\arraystretch}{1.1} 
\setlength{\tabcolsep}{6pt}
\begin{tabular}{l|cccc}
\toprule
\multirow{2}{*}{\textbf{Methods}}   & \multicolumn{4}{c}{\textbf{Jump}}                     \\ 
\cmidrule{2-5} 
                           & MELV   & MERP   & MEKB   & MLE               \\ 
\midrule
Classicial OT                 & 12.577  & 41.678 & 31.922 & 1458.255          \\
Order-preserving OT           & \textbf{14.294} & \textbf{42.063} & \textbf{32.865} & \textbf{1458.304}  \\
\midrule
1 generated samples              & 12.960  & 41.716 & 31.897 & 1458.275 \\
3 generated samples & 13.376 & 41.703 & 31.700 & 1458.245 \\
6 generated samples  & \textbf{14.294} & \textbf{42.063} & \textbf{32.865} & \textbf{1458.304} \\ 
\bottomrule
\end{tabular}
\caption{Ablation study results on Jump, evaluating with MELV, MERP, MEKB, and MLE metrics.}
\label{tab:abl}
\end{table}
%


\section{Conclusion}
\label{sec:conclusion}

\label{sec:concl}
We address the challenge of adapting robust humanoid motion policies using readily available walking motions, a single non-walking target sample, and a walking-trained base model. Our approach employs order-preserving optimal transport to synthesize meaningful intermediate poses along geodesic paths that smoothly bridge the source and target motions, and then refines these poses through collision-free optimization and retargeting to ensure physical plausibility and compatibility with the humanoid model. The resulting motion set provides an effective curriculum for reinforcement learning, enabling stable policy adaptation despite extreme target-motion data scarcity. Experiments show that our method consistently outperforms the baselines across tasks and evaluation settings, validating its ability to improve both generalization and performance while keeping target-motion data requirements minimal.

\section*{Acknowledgements}
Authors appreciate the support provided by the NYUAD Center for Artificial Intelligence and Robotics (CAIR), funded by Tamkeen under the NYUAD Research Institute Award CG010.

{
    \small
    \bibliographystyle{ieeenat_fullname}
    \bibliography{main}
}

\clearpage
\setcounter{page}{1}
\maketitlesupplementary




\noindent \textit{Collision detection} plays a pivotal role in generating physically plausible poses for articulated structures, where self-intersections may occur due to intricate joint arrangements and complex motion patterns. The process entails representing the skeleton as a hierarchical tree of joints and bones, transforming local joint rotations to global coordinates through forward kinematics, and assessing collisions via geometric primitives like spheres and capsules. Intersections are quantified using penetration energies, which are subsequently minimized through optimization to yield collision-free and physically consistent configurations. This framework draws on computational geometry and optimization techniques, ensuring differentiability for an efficient gradient-based solution.

\subsection{Forward Kinematics: Transforming Local Joint Rotations to Global Coordinates}
\label{subsec:forward}

The first step of collision detection is the computation of global joint positions from local joint rotations. Consider a skeleton with $J$ joints, governed by a parent array $\mathbf{p} \in \mathbb{Z}^J$, where $p_j = -1$ denotes the root joint, and otherwise indicates the parent of joint $j$. Each joint $j$ is parameterized by a local rotation quaternion $\mathbf{q}_j \in \mathbb{S}^3$ $\mathbf{q}_j \in \mathbb{H}_1 \subset \mathbb{R}^4$ represents the rotation of the $j$-th joint. Here, $\mathbb{H}_1 = \{ \mathbf{q} \in \mathbb{R}^4 : \|\mathbf{q}\|_2 = 1 \}$ denotes the unit quaternion group and a local translation $\mathbf{t}_j \in \mathbb{R}^3$ specified by the kinematic tree of the given robot.

Global transformations are derived recursively. The local transformation matrix is $\mathbf{T}_j^{\text{local}} = \begin{pmatrix} \mathbf{R}(\mathbf{q}_j) & \mathbf{t}_j \\ \mathbf{0}^\top & 1 \end{pmatrix}$, with $\mathbf{R}(\mathbf{q}_j)$ the rotation matrix induced by $\mathbf{q}_j$. The global matrix is $\mathbf{T}_j = \mathbf{T}_{p_j} \cdot \mathbf{T}_j^{\text{local}}$ for $p_j \neq -1$, and $\mathbf{T}_j = \mathbf{T}_j^{\text{local}}$ at the root. The global position $\mathbf{x}_j \in \mathbb{R}^3$ extracts the translation from $\mathbf{T}_j$. This yields the set $\{\mathbf{x}_j\}_{j=1}^J$, facilitating subsequent geometric evaluations.

\subsection{Geometric Primitives for Collision Modeling}
\label{subsec:primitive}
\begin{figure*}[!htb]
\centering
\includegraphics[width=.5\linewidth]{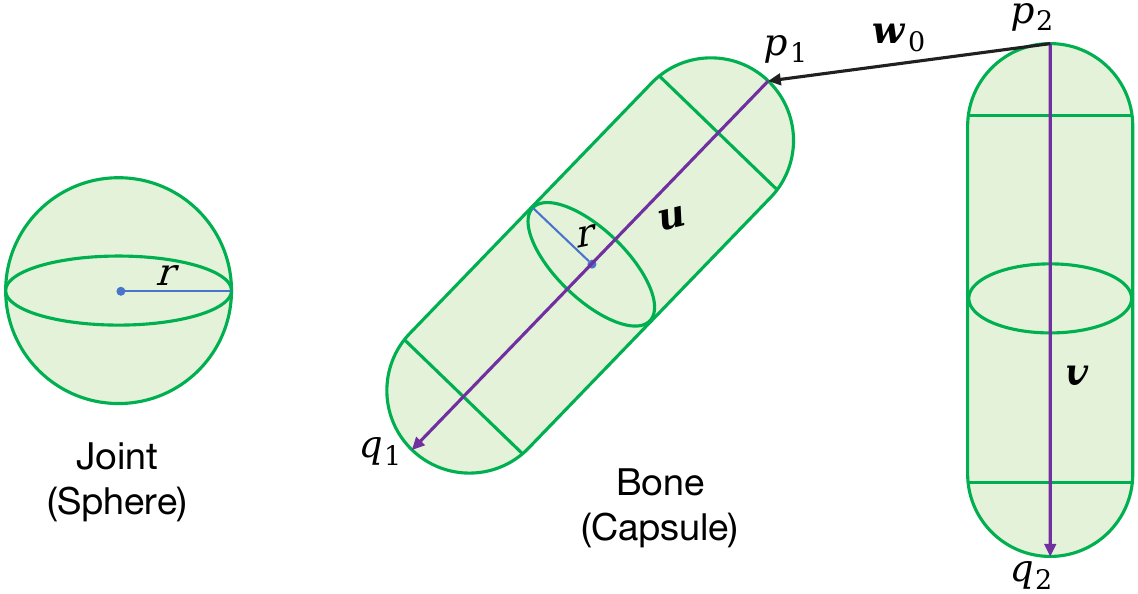}
\caption{Illustration of spheres for joints, capsules for bones, and line segment distance between two bones.}
\label{fig:joint_bone}
\end{figure*}

As shown in Fig.~\ref{fig:joint_bone}, joints are approximated as sphere with radii $r_j$, scaled by a factor $\rho$ (typically $\rho = 0.04$) proportional to bone lengths: $r_j = \rho \|\mathbf{t}_j\|$ for the root, or $r_j = \rho \|\mathbf{x}_j - \mathbf{x}_{p_j}\|$ otherwise, where $p_j$ represents the parent joint of the $j$th joint.

Bones, linking parent-child joints, are modeled as capsules---cylinders capped by hemispheres---with radius equal to the parent's sphere radius. A bone from joint $i$ to $j$ comprises the line segment $[\mathbf{p}, \mathbf{q}]$ where $\mathbf{p} = \mathbf{x}_i$, $\mathbf{q} = \mathbf{x}_j$, and radius $r_i$.

%

\subsection{Distance Computation for Line Segments}
\label{subsec:line_seg}

Capsule intersections necessitate the minimum distance between bone line segments, a core problem in computational geometry with roots in robotics and graphics. Here, we provide a way to compute the minimum distance between bone line segments~\cite{ericson2004real} (\ie, the purple lines in Fig.~\ref{fig:joint_bone}), not the bone capsule. For segments $[\mathbf{p}_1, \mathbf{q}_1]$ and $[\mathbf{p}_2, \mathbf{q}_2]$, the distance $d = \min \| \mathbf{x} - \mathbf{y} \|$ is sought, with $\mathbf{x}, \mathbf{y}$ are points on respective segments.

Parametrize: $\mathbf{x} = \mathbf{p}_1 + s \mathbf{u}$, $\mathbf{y} = \mathbf{p}_2 + t \mathbf{v}$, $s, t \in [0, 1]$, $\mathbf{u} = \mathbf{q}_1 - \mathbf{p}_1$, $\mathbf{v} = \mathbf{q}_2 - \mathbf{p}_2$. Minimize $f(s, t) = \| \mathbf{x} - \mathbf{y} \| = \| \mathbf{w}_0 + s \mathbf{u} - t \mathbf{v} \|^2$, $\mathbf{w}_0 = \mathbf{p}_1 - \mathbf{p}_2$. Geometrically, configurations include intersecting (distance zero), skew (unique common perpendicular), parallel (constant separation), and degenerate (collinear) cases. Closest points categorize as interior-interior, interior-vertex, or vertex-vertex.

Expanding $f(s,t)$ (using the dot product):
\begin{equation}
\begin{aligned}
& f(s,t) = (\mathbf{w}_0 + s\mathbf{u} - t\mathbf{v}) \cdot (\mathbf{w}_0 + s\mathbf{u} - t\mathbf{v}) \\
&= |\mathbf{w}_0|^2 + s^2 |\mathbf{u}|^2 + t^2 |\mathbf{v}|^2 + 2s \mathbf{w}_0 \cdot \mathbf{u} - 2t \mathbf{w}_0 \cdot \mathbf{v} - 2st \mathbf{u} \cdot \mathbf{v}\enspace.
\end{aligned}
\label{eq:dist}
\end{equation}
This is a quadratic function in $s$ and $t$, and since the Hessian (second derivatives) is positive semi-definite (as we'll see via the determinant), it has a global minimum.
\noindent \textbf{Deriving the linear system via critical points.}
To find the minimum, compute the partial derivatives of $f(s, t)$ and set them to zero (stationarity conditions from calculus):
\begin{equation}
\frac{\partial f}{\partial s} = 2s |\mathbf{u}|^2 + 2 \mathbf{w}_0 \cdot \mathbf{u} - 2t \mathbf{u} \cdot \mathbf{v} = 0\enspace,
\end{equation}
\begin{equation*}
\frac{\partial f}{\partial t} = 2t |\mathbf{v}|^2 - 2 \mathbf{w}_0 \cdot \mathbf{v} - 2s \mathbf{u} \cdot \mathbf{v} = 0\enspace.
\end{equation*}
Dividing both by 2 and rearranging:
\begin{equation}
s |\mathbf{u}|^2 - t (\mathbf{u} \cdot \mathbf{v}) = - (\mathbf{w}_0 \cdot \mathbf{u})\enspace,
\end{equation}
\begin{equation}
-s (\mathbf{u} \cdot \mathbf{v}) + t |\mathbf{v}|^2 = \mathbf{w}_0 \cdot \mathbf{v}\enspace.
\end{equation}
Define the scalars:
\begin{itemize}
\item $a = \mathbf{u} \cdot \mathbf{u} = |\mathbf{u}|^2$,
\item $b = \mathbf{u} \cdot \mathbf{v}$,
\item $c = \mathbf{v} \cdot \mathbf{v} = |\mathbf{v}|^2$,
\item $d = \mathbf{u} \cdot \mathbf{w}_0$,
\item $e = \mathbf{v} \cdot \mathbf{w}_0$.
\end{itemize}
This gives the linear system in matrix form:
\begin{equation}
\begin{pmatrix}
a & -b \\
b & -c
\end{pmatrix}
\begin{pmatrix}
s \\
t
\end{pmatrix}
\begin{pmatrix}
-d \\
-e
\end{pmatrix}.
\end{equation}
Geometrically, these equations enforce that the vector connecting the closest points, $\mathbf{x}(s) - \mathbf{y}(t)$, is perpendicular to both direction vectors $\mathbf{u}$ and $\mathbf{v}$. This is because:
\begin{equation}
\frac{\partial f}{\partial s} = 2(\mathbf{x}(s) - \mathbf{y}(t)) \cdot \mathbf{u} = 0 \implies (\mathbf{x}(s) - \mathbf{y}(t)) \perp \mathbf{u}\enspace,
\end{equation}
\begin{equation}
\frac{\partial f}{\partial t} = -2(\mathbf{x}(s) - \mathbf{y}(t)) \cdot \mathbf{v} = 0 \implies (\mathbf{x}(s) - \mathbf{y}(t)) \perp \mathbf{v}\enspace.
\end{equation}
For non-parallel lines, this common perpendicular is unique, leading to a unique solution $(s, t)$.

\noindent \textbf{The determinant $D$ and its properties.}
The determinant of the coefficient matrix is:
\begin{equation}
\det = a(-c) - (-b)b = -ac + b^2 = -(ac - b^2)\enspace.
\end{equation}
By the Cauchy-Schwarz inequality:
\begin{equation}
|\mathbf{u} \cdot \mathbf{v}|^2 \leq |\mathbf{u}|^2 |\mathbf{v}|^2 \implies b^2 \leq ac \implies D = ac - b^2 \geq 0\enspace.
\end{equation}
\begin{itemize}
\item Strict inequality $D > 0$: Holds when $\mathbf{u}$ and $\mathbf{v}$ are not parallel (\ie, the lines are skew or intersecting but not parallel). The system has a unique solution.
\item Equality $D = 0$: Occurs when $\mathbf{u}$ and $\mathbf{v}$ are parallel ($\mathbf{u} = k \mathbf{v}$ for some scalar $k \neq 0$). The lines are parallel, and the system may be consistent (infinite solutions if the separation is constant) or inconsistent (no solution if non-coplanar, but in distance minimization, we handle via projections). In practice, for $D \approx 0$, we treat it as a special case to avoid division by zero.
\end{itemize}
The non-negativity property of $D$ ensures the quadratic form is convex, guaranteeing a minimum (or saddle in degenerate cases).

\noindent \textbf{Solving the system for $D > 0$.}
For the non-parallel case ($D > 0$), solve using Cramer's rule or matrix inversion. The solutions are:
\begin{equation}
\begin{aligned}
s &= \frac{\det \begin{bmatrix} -d & -b \\ -e & -c \end{bmatrix}}{-D} = \frac{(-d)(-c) - (-b)(-e)}{-D} \\
&= \frac{dc - be}{-D} = \frac{be - cd}{D}\enspace, \\
t &= \frac{\det \begin{bmatrix} a & -d \\ b & -e \end{bmatrix}}{-D} = \frac{a(-e) - (-d)b}{-D} \\
&= \frac{-ae + db}{-D} = \frac{bd - ae}{D} = \frac{ae - bd}{D}\enspace.
\end{aligned}
\end{equation}
For finite segments, we clamp $s, t \in [0,1]$. Then, we re-adjust these two values, \ie, fix $t$, solve $s = (d + t b)/a$, re-clamp into $[0,1]$ again. After getting $s$ and $t$, we plug them into Eq.~\ref{eq:dist} to compute the minimum distance between two line segments. 

\noindent \textbf{Handling the parallel case ($D = 0$).}
When $D \approx 0$, \ie, $D < \epsilon$ where $\epsilon \approx 10^{-9}$, the lines are parallel, and the shortest distance is the perpendicular separation between the lines (constant along their length). There is no unique $(s, t)$; instead:
\begin{itemize}
\item Project $\mathbf{w}_0$ onto the common direction (normalized $\mathbf{v} / |\mathbf{v}|$).
\item Choose an arbitrary $s$ (\eg, $s = 0$), then solve for $t = e / c$ (clamped if finite), or compare endpoint projections.
\item The minimum distance is $|\mathbf{w}_0 - \text{proj}_{\mathbf{v}} \mathbf{w}_0|$. This ensures numerical stability and geometric correctness..
\end{itemize}
For finite segments, the distance is computed in $O(1)$ time, robust for real-time applications.

\subsection{Penetration Energies: Quantifying Collisions}
\label{subsec:energy}
Energies penalize intersections over non-adjacent pairs, avoiding kinematic artifacts.

For (joint) spheres, the energy $E_{\text{sphere}}$ is the sum of squared penetrations:
\begin{equation}
E_{\text{sphere}} = \sum_{i<j,(i,j)\in\mathcal{V}} \left( \max(0, r_i + r_j - |\mathbf{x}_i - \mathbf{x}_j|) \right)^2\enspace,
\end{equation}
where $\mathcal{V}$ excludes adjacent or grandparent-grandchild pairs (\ie, $j \neq p_i$, $i \neq p_j$, \etc).

For (bone) capsules, gather bones as pairs $(a_m, b_m)$ for $m = 1, \dots, M$ (typically $M = J-1$). The energy $E_{\text{capsule}}$ is defined as:
\begin{equation}
E_{\text{capsule}} = \sum_{i<j,(i,j)\in\mathcal{V}'} \left( \max(0, r_i + r_j - d_{ij}) \right)^2\enspace,
\end{equation}
where $d_{ij}$ is the line segment distance between bones $i$ and $j$, \ie, the minimum value of Eq.~\ref{eq:dist} as discuss above, and $\mathcal{V}'$ excludes bones sharing joints.

The total energy is $E = E_{\text{sphere}} + \lambda E_{\text{capsule}}$ where we set $\lambda = 1$ in our settings.

\subsection{Pose Optimization for Collision Resolution}
\label{subsec:opt}
The total energy $E$ is differentiable \wrt the joint rotations. Thus, we minimize $E$ over $\{\mathbf{q}_j\}$ on $\text{SO(3)}^J$ using Riemannian gradient descent (learning rate $\eta \approx 0.05$, \eg, 120 steps). Project gradient: $\mathbf{g}_{\text{proj}} = \nabla_{\mathbf{q}} E - (\mathbf{q} \cdot \nabla_{\mathbf{q}} E) \mathbf{q}$. Update $\mathbf{q} \leftarrow \mathbf{q} - \eta \mathbf{g}_{\text{proj}}$, and then normalize, \ie, send back to $\text{SO(3)}^J$. This process is sketched in Algorithm 1 in the main text. This iteratively mitigates penetrations, producing viable poses for animation and simulation.


\end{document}